\documentclass[10pt,twocolumn,letterpaper]{article}

\usepackage{cvpr}              % To produce the CAMERA-READY version

\usepackage{comment}

\usepackage{colortbl} 
\usepackage[dvipsnames]{xcolor}

\newcommand{\modelname}{Mirror-Aware Neural Humans}

\newcommand{\DA}[1]{{}}
\newcommand{\da}[1]{{#1}}

\newcommand{\danred}[1]{{#1}}

\newcommand{\dotcaption}[2]{%
  \caption[#1]{\textbf{#1}. {#2}}%
}

\newcommand{\shortcaption}[2]{%
  \caption[#1]{\textbf{#1} {#2}}%
}

\renewcommand{\comment}[1]{}
\newcommand{\parag}[1]{\noindent\textbf{#1}}

\newcommand{\vb}{\mathbf{b}}
\newcommand{\vc}{\mathbf{c}}
\newcommand{\vd}{\mathbf{d}}

\newcommand{\vell}{\boldsymbol{\ell}}
\newcommand{\vf}{\mathbf{f}}

\newcommand{\vm}{\mathbf{m}}
\newcommand{\vn}{\mathbf{n}}

\newcommand{\vp}{\mathbf{p}}
\newcommand{\vq}{\mathbf{q}}
\newcommand{\vr}{\mathbf{r}}
\renewcommand{\vs}{\mathbf{s}}

\newcommand{\vtheta}{\boldsymbol{\theta}}

\newcommand{\vv}{\mathbf{v}}

\newcommand{\mA}{\mathbf{A}}

\newcommand{\mI}{\mathbf{I}}

\newcommand{\mK}{\mathbf{K}}
\newcommand{\mL}{\mathbf{L}}
\newcommand{\mM}{\mathbf{M}}
\newcommand{\mN}{\mathbf{N}}

\newcommand{\mR}{\mathbf{R}}

\newcommand{\mT}{\mathbf{T}}

\newcommand{\cL}{\mathcal L}

\newcommand{\cN}{\mathcal N}

\newcommand{\Real}{\mathbb{R}}

\definecolor{cvprblue}{rgb}{0.21,0.49,0.74}
\usepackage[pagebackref,breaklinks,colorlinks,citecolor=cvprblue]{hyperref}

\def\paperID{12} % *** Enter the Paper ID here
\def\confName{3DV\xspace}
\def\confYear{2024\xspace}

\title{Mirror-Aware Neural Humans}

\author {Daniel Ajisafe$^1$ \qquad James Tang$^1$ \qquad Shih-Yang Su$^1$ \qquad  Bastian Wandt$^{1,2}$ \qquad Helge Rhodin$^1$ \\
{\tt\small (dajisafe,shihyang,rhodin)@cs.ubc.ca, tangytob@student.ubc.ca, bastian.wandt@liu.se} \\
$^1$The University of British Columbia \\
$^2$Linköping University
}

\begin{document}
\maketitle
\begin{abstract}

Human motion capture either requires multi-camera systems or is unreliable when using single-view input due to depth ambiguities. Meanwhile, mirrors are readily available in urban environments and form an affordable alternative by recording two views with only a single camera. However, the mirror setting poses the additional challenge of handling occlusions of real and mirror image. Going beyond existing mirror approaches for 3D human pose estimation, we utilize mirrors for learning a complete body model, including shape and dense appearance. Our main contributions are extending articulated neural radiance fields to include a notion of a mirror, making it sample-efficient over potential occlusion regions. Together, our contributions realize a consumer-level 3D motion capture system that starts from off-the-shelf 2D poses by automatically calibrating the camera, estimating mirror orientation, and subsequently lifting 2D keypoint detections to 3D skeleton pose that is used to condition the mirror-aware NeRF. We empirically demonstrate the benefit of learning a body model and accounting for occlusion in challenging mirror scenes.  The project is available at: \url{https://danielajisafe.github.io/mirror-aware-neural-humans/}.

\end{abstract}    
\section{Introduction}
\label{sec:intro}

Estimating detailed 3D geometry of a moving person from a single video is a long-standing goal.
Learning-based solutions can succeed when trained on 3D labels from the target domain or when multiple 2D views are available for supervision~\cite{rhodin2018learning,rhodin2018unsupervised, rhodin2019neural,Unsup3DPose,mitra2020multiview,viewinvariant3DPose, wandt2021canonpose}. However, multi-view capture is expensive and tedious to calibrate, and hence, the diversity of existing datasets and associated machine learning solutions are limited to mainstream activities and environments.

\begin{figure}[t]%
\begin{center}
\includegraphics[width=1\linewidth,trim={0cm 0 0cm 0},clip]{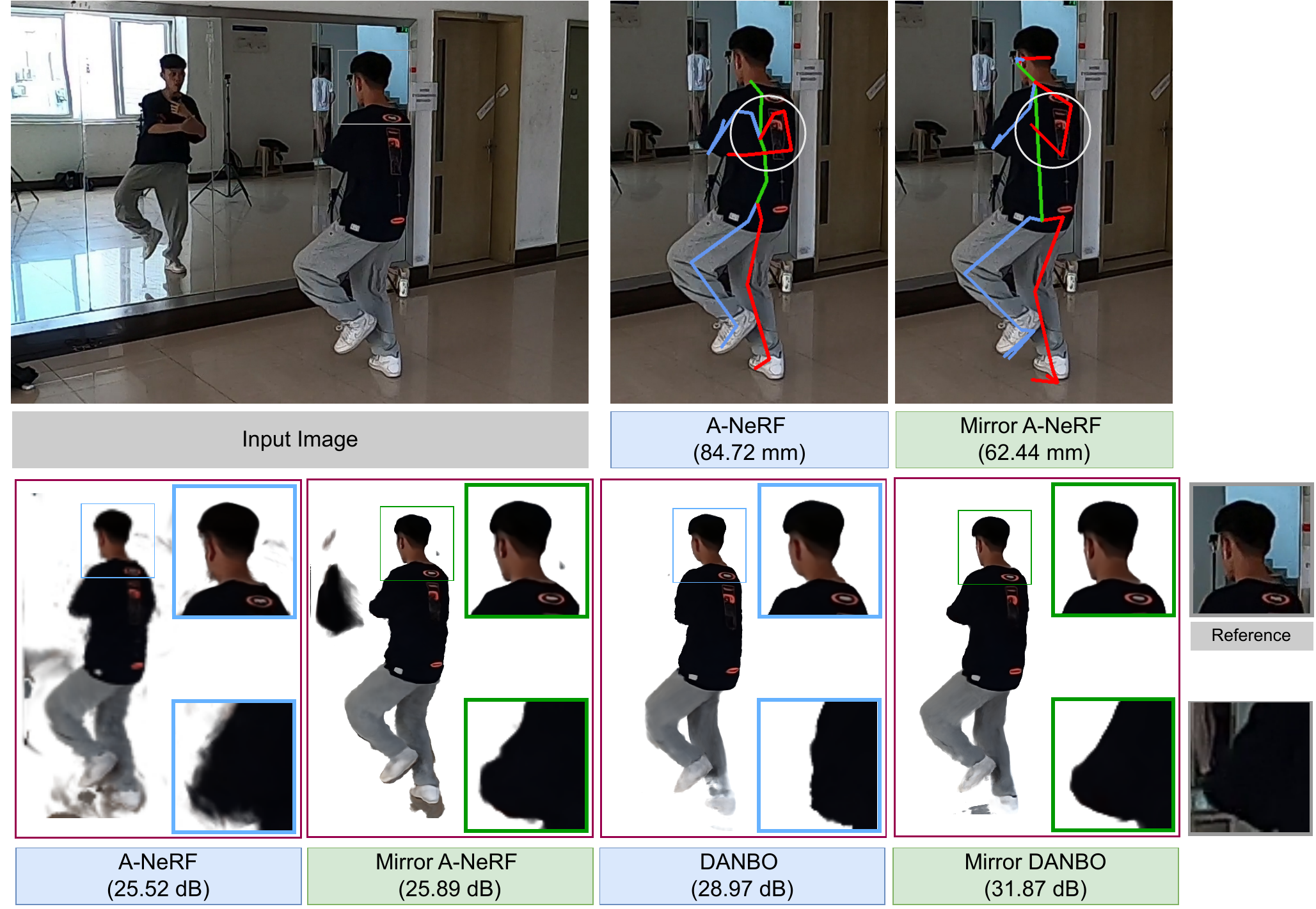}
\end{center}
  \dotcaption{Pose refinement and image quality}{Given an image with mirror (top left) our mirror-based method reconstructs 3D pose and shape that is more accurate than the baselines (A-NeRF~\cite{su2021nerf} and DANBO~\cite{su2022danbo}) not supporting the mirror, both in terms of the 3D pose metric PA-MPJPE (top row, e.g., corrected arms), and in image quality PSNR (bottom row, e.g., reconstructed earphone and left elbow).
  % \da{on evaluation set.}
  }
\label{fig:top_figure} 
\end{figure}

We propose a test-time optimization method for reconstructing a generative body model entailing pose, shape, and appearance using a single camera and a mirror and starting from 2D pose without any 3D labels nor large-scale dataset. The mirror setting is practical: 
First, mirrors are readily available in urban environments and provide a second view for accurate reconstruction without requiring multi-camera recording and temporal synchronization.
Second, off-the-shelf 2D estimators generalize well since diverse training images are easily annotated by clicking 2D joint locations.  Previous works \cite{fang2021reconstructing,liulearning} leveraged reflections in mirrors for better human pose reconstruction. However, neither of them model shape and appearance in detail. In addition, the model proposed in~\cite{fang2021reconstructing} needs a 3D pose estimator as prior,
%as a starting point 
potentially limiting their approach to 
%previously captured 3D motions.
motions close to the training set.

Alternatively, 3D body models have been learned from monocular video. However,
%is difficult from a single video, since it is high dimensional and underconstrained. 
existing approaches either use a shape prior, such as a scan of the person~\cite{habermann2019livecap}, a parametric body model~\cite{alldieck2019learning}, restrict motions to be simple~\cite{yang2021banmo}, or require initialization with a prior 3D
% n off-the-shelf 3D pose
estimator~\cite{su2021nerf, su2022danbo, weng2022humannerf}. This restricts the possible motion, shape, and appearance complexity. By contrast, our mirror setting is simple, enabling anyone to collect 3D data of their target domain.

Our approach for learning \textit{\modelname}
makes no prior assumption about the body shape by building upon
the open-source articulated neural radiance fields (NeRF) 
%(A-NeRF) 
models~\cite{su2021nerf,su2022danbo}, which require only the joint angles of a skeleton as input. We estimate this skeleton fully automatically and without prior assumptions on the 3D poses using an automatic mirror calibration (Step 1) and mirror-based 2D to 3D pose lifting (Step 2), thereby avoiding the use of 3D pose estimators that struggle with occlusions and extreme poses.
Our core contribution are as follows:
\begin{itemize}[noitemsep]
\item Designing a robust algorithm that estimates mirror position and orientation, and 3D skeleton model with bone-relative coordinates suitable for neural body models.
\item A layered mirror model, extending NeRF with occlusion handling of the mirror image by the real person.

\item Developing a complete motion capture system for reconstructing human pose, shape, and appearance from mirror images, and making the source code available at: \url{https://github.com/danielajisafe/Mirror-Aware-Neural-Humans}.

\end{itemize}

\comment{
Besides developing a complete motion capture system, for which we will publish source code, our core contribution is to integrate a mirror model into the underlying neural radiance field (NeRF), including an efficient occlusion handling between the real and the mirrored person. Figure \ref{fig:top_figure} highlights the improvements gained by utilizing the mirror.
}

\comment{
Figure~\ref{fig:teaser} visualizes all the involved steps and representations. Our contributions within these steps are:

\begin{itemize}%[noitemsep]
\item \textbf{Step 1.} Designing a robust algorithm that estimates mirror position and orientation from only 2D human pose detections by using the person as a calibration object, without requiring any manual annotation or labeling.
\item \textbf{Step 2.} Reconstructing a kinematic skeleton model from mirror images without requiring any pre-trained 3D model or dataset by minimizing the reprojection of the 3D model on the 2D detections.
Different from other mirror-based approaches~\cite{liulearning,Flight}, we recover smooth bone orientations to enable bone-relative coordinates in the subsequent mirror-extended NeRF model.

\item \textbf{Step 3a.} Designing a \emph{mirror NeRF} to form a generative human body model that recovers a complete volumetric body from a single video without 3D scans.
\item \textbf{Step 3b.} Making the mirror-aware body model sample-efficient by sampling adaptively to possible occlusion areas.
\end{itemize}

Our experiments demonstrate that not only is a full 3D body model learned from a single 2D video without prior 3D knowledge, but that 3D pose refinement improves upon geometric 2D to 3D keypoint lifting.}

\section{Related Work}
\label{sec:Related Work}

\paragraph{Self- and weakly supervised learning approaches.}

%Some approaches
Weakly supervised 3D pose estimators typically leverage small-scale 3D pose 
%from available 
datasets and combine them with additional 2D data \cite{WanRos2019a,habibie2019wild,Wang_2019_ICCV,Kundu_2020_CVPR,zanfir20normalizing,chen20garnet,habekost20learning,wandt2022elepose}.
Others utilize neural networks that are pretrained on the 3D lifting task and transfer them to another dataset \cite{mehta2017monocular,VNect_SIGGRAPH2017,pavllo20193d,guan2021bilevel,gong2021poseaug,gholami2022adaptpose}.
Such weak supervision transfers better to unseen poses.
However, they still make the assumption that training poses are close to the training set.
A different approach is using multi-view %\mbox{(self-)}
supervision to learn an embedding of 3D poses \cite{rhodin2018learning,rhodin2018unsupervised, rhodin2019neural,Unsup3DPose,mitra2020multiview,viewinvariant3DPose}
%. Without the unsupervised embedding step, others 
or learn the 3D reconstruction step directly from multi-view images \cite{rochette2019weakly,kocabas2019epipolar,Iqbal_2020_CVPR,wandt2021canonpose}.
While promising, they still require multiple temporally synchronized cameras for training. 
In contrast, using mirrors in a scene gives the unique advantage of having a pair of synchronized views with a single recording device.
%for free. 

\parag{Mirror geometry and calibration.}
Mirrors have a long history in visual computing on which
Reshetouski et al.~\cite{reshetouski2013mirrors} provide a good overview.
%Several approaches 
We take inspiration from methods
~\cite{orthogonality,Stereo,RGBD,Hu2005MultipleView3R} employing mirrors for camera calibration and 3D reconstruction of rigid objects,
%We also automate extrinsic camera calibration, 
%and generalize them for moving humans instead of a rigid reference object.
to enable calibration and reconstruction of moving humans. Alternatively,  Yin et al.~\cite{yin2023multi} reconstructs arbitrary objects in mirror-like surfaces but do not show any application for humans.

\parag{Mirror-based human pose estimation.} 
Nguyen et al.~\cite{Flight} use mirrors to reconstruct human point clouds, but require a depth camera together with two or multiple mirrors. 
To the best of our knowledge, the most related work that reconstructs human pose and body shape with a single mirror is from Fang et al.~\cite{fang2021reconstructing}. 
They provide an optimization-based approach that utilizes mirror symmetry constraints for predicting 3D human pose and mirror orientation. 
While attaining high accuracy, they require as input an initial 3D pose estimate from a pretrained neural network that cannot generalize well to unseen poses. Moreover, their best results are attained using manually annotated vanishing lines on the mirror boundary~\cite{fang2021supp}. 
By contrast, we use a purely geometric approach to optimize for 3D keypoints without requiring any 3D pose estimator or mirror annotation (with the neural network only modeling shape and appearance), by jointly optimizing for the bone orientation and building upon recent work on estimating camera position and ground plane using the motion of people in the scene~\cite{tang2016camera,fei2021single}. Similar to prior approaches~\cite{fang2021reconstructing, liulearning}, 
% Fang et al.~\cite{fang2021reconstructing}, Liu et al.~\cite{liulearning}
we estimate 3D human keypoints as a solution to an optimization problem between two sets of mirrored 2D keypoints. 
By contrast, Liu et al.~\cite{liulearning} optimize for 3D joint coordinates which can lead to incorrect pose sequences where, for example, bone lengths vary over time, and orientation remains ambiguous. Fang et al.~\cite{fang2021reconstructing} restrict motions to be close to sequences previously captured by pre-trained detectors, and none of these methods take detailed reconstruction of shape and appearance into account.

\section{Method}
\label{ch:Method}

\begin{figure*}[t!]
\centering
\includegraphics[width=0.95\textwidth,trim={0cm 0cm 0cm 0cm},clip]{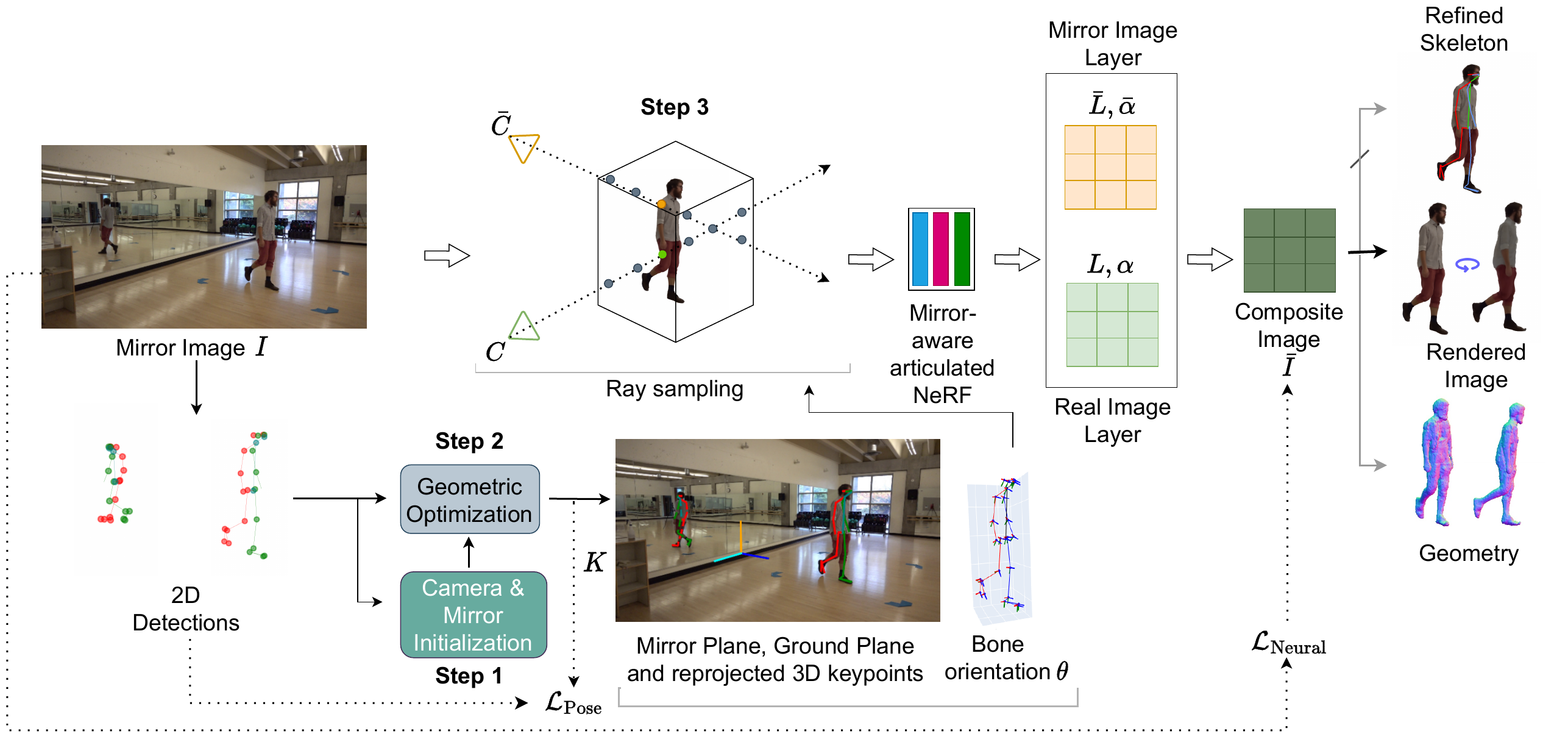}
  \caption{We start from a mirror image with an unknown mirror geometry. With only 2D detections and suitable assumptions, we reconstruct the mirror plane, ground plane, and 3D keypoints in Step 1 and Step 2. 
  %, without any ground truth. 
  Our optimization yields bone orientation that is crucial for
  %ray sampling and 
  integrating NeRF with the mirror-based reconstruction. 
  %\danteal{and awareness}
  %to get a \emph{mirror-aware articulated NeRF}. 
  The final Mirror-aware Neural Human is learned via layered composition of mirror and real images in Step 3 and yields improved body pose, %(optionally),
  shape, and appearance quality.
  }
\label{fig:overview} 
\end{figure*}

Our goal is to reconstruct a dense neural body model from a single video with only sparse 2D detections as input, using the mirror as a second view to impose multi-view constraints. The difficulty lies in reconstructing such dense representation from only sparse and noisy 2D labels, with an unknown mirror and camera configuration. %\danteal{This uncertainty also implies} 
By contrast to classical multi-view settings, mirror observations add the difficulty of the real person occluding the mirror image. To overcome these difficulties, our method goes from sparse to fine details in three steps, as sketched in Figure~\ref{fig:overview}.

\begin{figure}[t]%
\begin{center}
\includegraphics[width=1.0\linewidth,trim={0.0cm 0 3.5cm 0},clip]{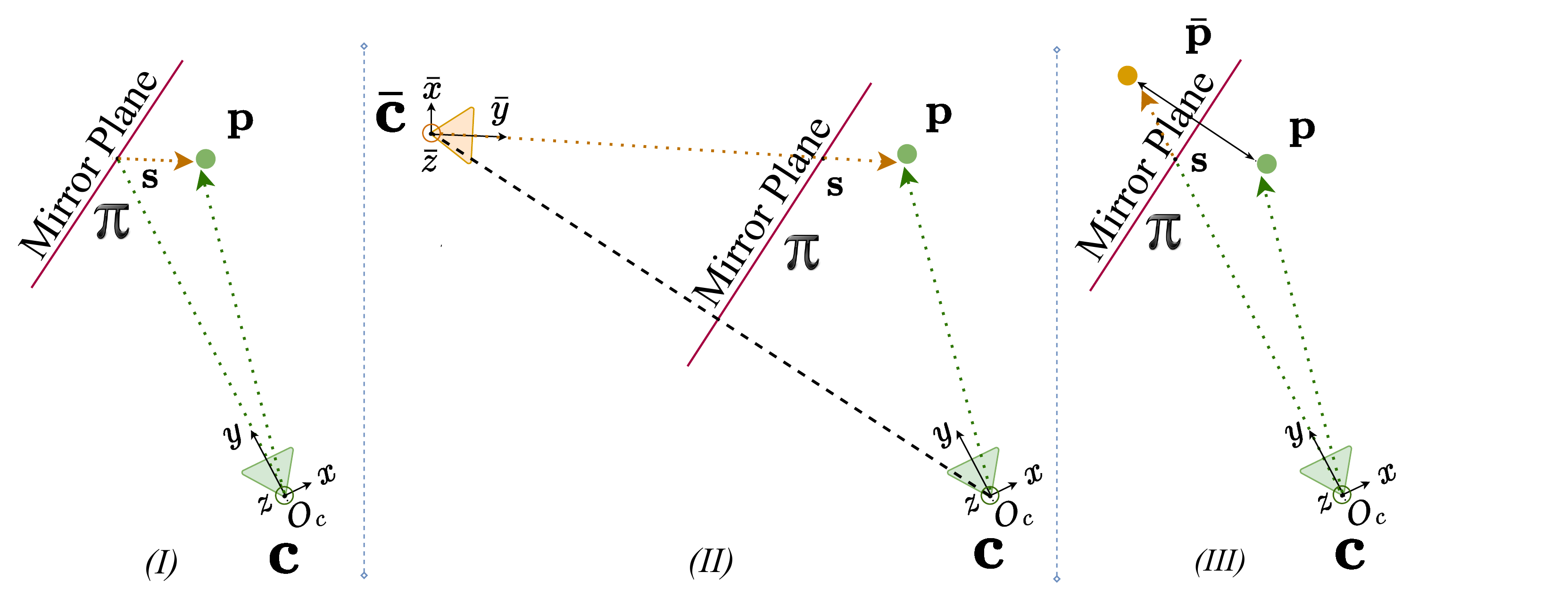}
\end{center}
\dotcaption{Models for the mirror reflection} {In the first case, the rays go from the real camera $\vc$ up to the mirror plane $\pi$ intersecting at location $\vs$, then to the real person $\vp$ after a mirror reflection. In the second case, the real person $\vp$ is viewed from a virtual camera $\bar{\vc}$ forming a virtual image. In the third case, the person location is mirrored to $\bar{\vp}$ and light rays go straight from camera $\vc$ to $\bar{\vp}$.}
\label{fig:mirror-geo}  
\end{figure}

For each step we use a suitable representation for the mirror geometry, each mathematically equivalent yet implying a different implementation. Figure \ref{fig:mirror-geo} visualizes the three forms. 
%There are three equivalent ways of modeling the reflection. 
\emph{Case~I:} A single camera $\vc$ with light rays reflecting on the mirror plane $\pi$. \emph{Case~II:} The mirror image stemming from a \emph{virtual camera} $\bar{\vc}$ opposing the real camera $\vc$. \emph{Case~III:} A \emph{virtual person} $\bar{\vp}$ opposing the real person $\vp$, both viewed from the real camera $\vc$. 


\subsection {Camera and Mirror Initialization (Step 1)}
\label{ssec:mirror calib}
We start from a video that shows a person moving in front of a mirror and use off-the-shelf pose detectors 
% \cite{liu2021deep, xu2022vitpose}
\cite{alphapose, xu2022vitpose} to obtain 2D pose estimates 
$\vq^{(t)}\in \Real^{2\times J}$ 
for every input frame $\mI_t$ and all $J$ joints. As we assume the mirror is orthogonal to the ground, mirror and real images appear to be standing on the same ground plane and existing solutions to using the human as calibration object apply. 
%As no open source solution is available, we implemented a variant of 
We use a variant of \da{single-view distance estimation}~\cite{fei2021single} as described in \da{method}~\cite{tang2023anon} that yields focal length $f$ and ground plane normal $n_g$. 
%\danteal{Based on this, with only two or three detected poses and perspective projection we reconstruct the groundplane normal $n_g$, details are in the supplemental.}

\begin{figure}[t]
\begin{center}
\includegraphics[width=1.0\linewidth,trim={11cm 0 0 0cm},clip]{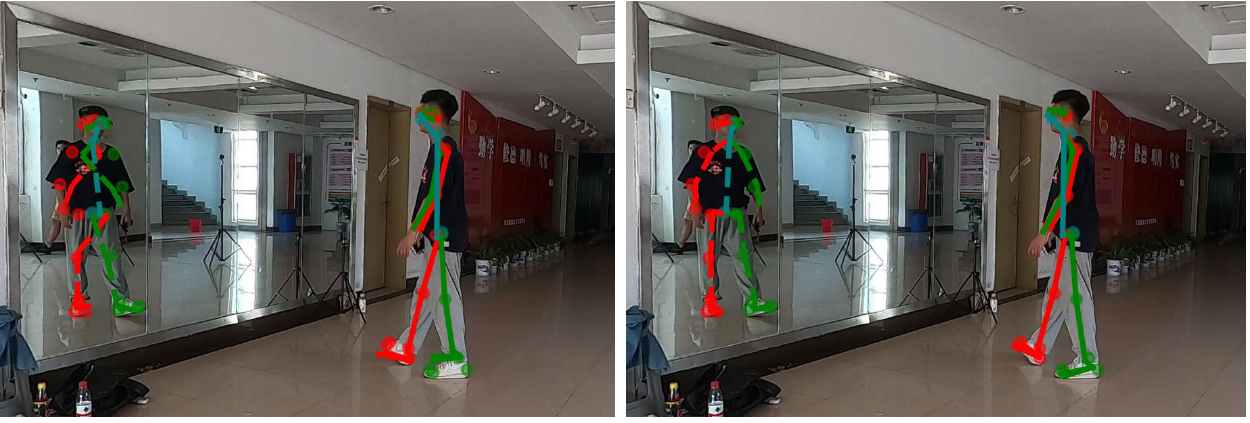}
\end{center}
  \caption{\textbf{Real and mirror pose assignment.} Our algorithm
distinguishes the real from the virtual person using pelvis-to-neck distance. 
%The skeleton with the highest distance is \dan{assigned} the real person. 
With the right assignment \da{and correct flipping}, cases of collapsed poses (left)  are corrected (right).
}
\label{fig:flip_and_J_collapse} 
\end{figure}

\paragraph{Associating real and mirror poses.} Pose detectors are not aware of the mirror and therefore treat each person independently. We associate the pose with the largest neck-to-pelvis distance as the real person, utilizing that the person viewed through the mirror is farther away and hence smaller in the perspective camera. This association is also required for flipping the left and right side of the mirrored 2D pose to account for the mirror operation. Figure \ref{fig:flip_and_J_collapse} shows this relationship and the degradation when \da{misaligned}.

\paragraph{Mirror geometry initialization.} 

Under the assumption that the mirror normal is orthogonal to the ground plane normal, we obtain the 3D location of the real person and mirrored person using \emph{Case~III} (see Figure~\ref{fig:mirror-geo}). 
We project their 2D ankle locations $\vq_\text{ankle}$ onto
the estimated ground plane by reversing the projection
    \begin{equation}
    \vq = \mK \vp,
    \text{ where }
\mK = 
\begin{pmatrix}
f & 0 & o_1 & 0\\
0 & f & o_2 & 0\\
0 & 0 & 1 & 0
\end{pmatrix},
\label{eq:projection}
\end{equation}
with $(o_1,o_2)$ the image center and $f$ the estimated focal length. The  mirror normal $\vn_{m} \in \Real^{3}$ 
is then the vector from real $\vp_\text{ankle}$ to mirror $\bar{\vp}_\text{ankle}$,
\begin{equation}
    \vn_m = \frac{\vp_\text{ankle}-\bar{\vp}_\text{ankle}}{\|\vp_\text{ankle}-\bar{\vp}_\text{ankle}\|}.
\end{equation} 
The mirror location is the midpoint, $\vm = (\vp_\text{ankle}-\bar{\vp}_\text{ankle})/2$. For increased robustness, we average over all frames.

\subsection{2D to 3D Pose Lifting (Step 2)} \label{ssec:poselift}

In this section, we use the notion of a virtual camera (\emph{Case~II})
positioned behind the mirror
as shown in Figure~\ref{fig:mirror-geo}.
Following~\cite{rotation_arbitrary, schwertz2012field}, we derive the virtual camera through the matrix $\mA$ that mirrors points across the mirror plane, 
\begin{equation}
\mA =
\begin{bmatrix}
1-2n_{x}^{2} & -2n_{y}n_{x} & -2n_{z}n_{x} & -2n_{x}d\\
-2n_{y}n_{x} & 1-2n_{y}^{2} & -2n_{y}n_{z} & -2n_{y}d\\
-2n_{z}n_{x} & -2n_{y}n_{z} & 1-2n_{z}^{2} & -2n_{z}d\\
0 & 0 & 0 & 1
\end{bmatrix},
\end{equation}
with $\vn_m = [n_x,n_y,n_z]$ the mirror normal and $d$ the distance between camera and mirror. Both quantities are from Step~1. By defining the real camera to be at the origin pointing along the z-axis, $\mA$ maps points from the real to the virtual camera. The orientation of the virtual camera is hence $\bar{\mR} = \mA_{3\times3}^{\top}$, the inverse of the top-left part of $\mA$, and camera position $\bar{c} = -2\vn_m d$, is the negative of the last column of $\mA$.
Note that $\bar{\mR}$ is from the orthogonal group $O(3)$ %but not from $SO(3)$ 
as it includes a reflection component given by the mirror.

\paragraph{Mirror skeleton representation.}
To be able to reconstruct not only the position but also the orientation of limbs, we represent $\vp^{(t)}$ with a skeleton parameterized by joint rotations $\vtheta^{(t)}_i \in \Real^{6}$, using the 6D rotation parameterization of \cite{zhou2019continuity}, bone lengths $\vell \in \Real^{J}$, and the 3D pelvis position $\vp_\text{pelvis}^{(t)} \in \Real^{3}$ (the root position). Forward kinematics gives
\begin{gather} 
\vp^{(t)}_j = %\left(
\prod_{i \in \cN(j)} \mT_i%\right)
%\vc{c{p}^\text{ref}_j  
\begin{bmatrix}
\bf{0}\\1
\end{bmatrix}
+\vp_\text{pelvis}^{(t)}, \mT_i = 
\begin{bmatrix} \mM(\vtheta^{(t)}_{i})& {\vell}_{i}\vv_i\\ \mathbf{0} &  {1} \end{bmatrix},
\end{gather}
with $\vv^\text{ref}_i \in \Real^{3}$ the $i$th bone vector (parent to child vector) in a reference pose, $\mM(\vtheta^{(t)}_{i})$ the joint rotation computed from $\vtheta^{(t)}_{i}$, and $\cN(j)$ the ancestors of $j$ in the kinematic chain. 
In the following, we optimize these parameters by using different constraints on the mirror scene including the pose, feet, and bone orientation.

\begin{figure}[t]
\centering
\includegraphics[width=1.0\linewidth,trim={0.5cm 0.5cm 0.5cm 0.5cm},clip]{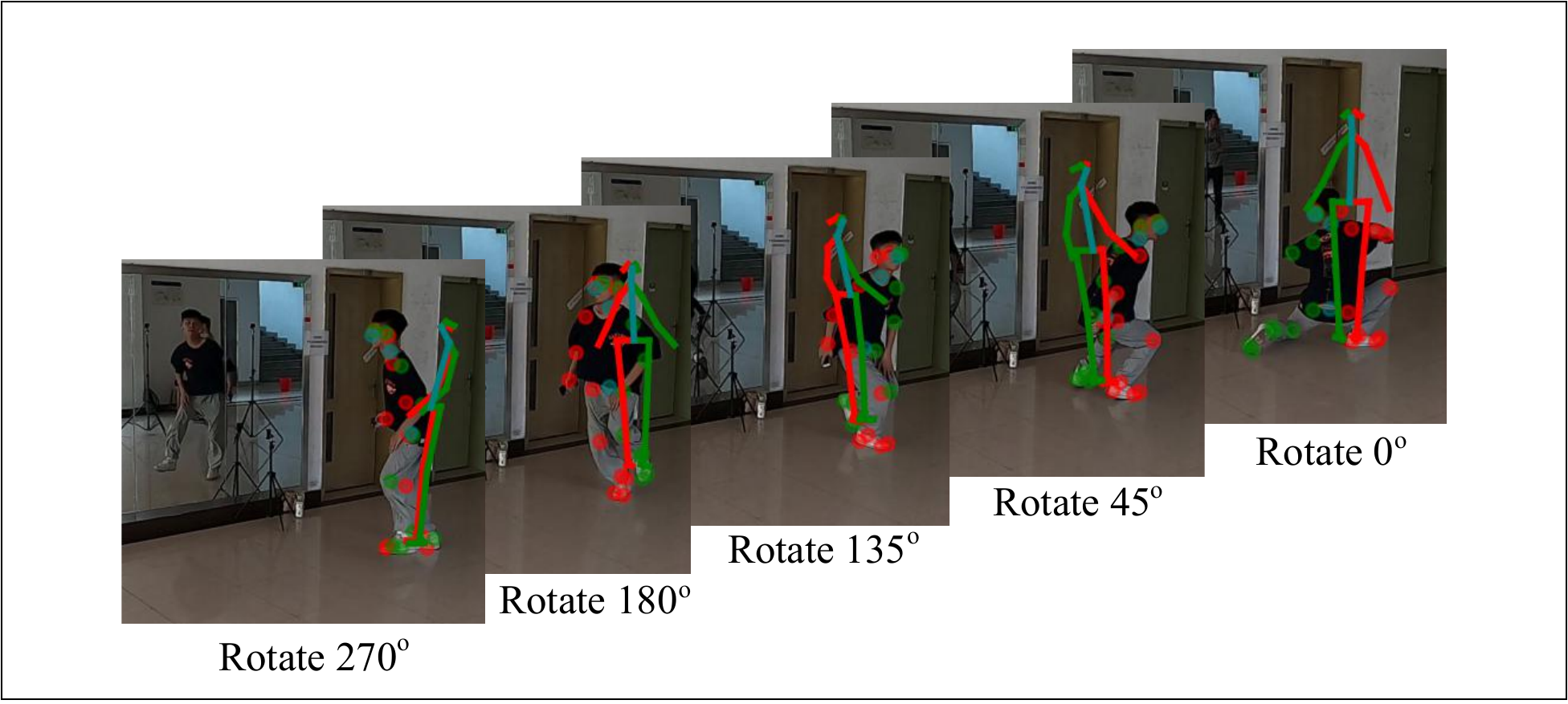}
  \dotcaption{3D pose initialization} {by measuring the error between initial re-projections (lines forming skeleton) and 2D detections (dots) to determine the optimal starting pose.
  }
\label{fig:initialization} 
\end{figure}

\paragraph{3D pose initialization.} Unlike prior work using joint locations~\cite{liulearning}, the bone rotation estimation we require is prone to local minima. To overcome this, we initialize with a constant standing pose adapted from the first frame of H36M~\cite{ionescu2013human3, ionescu2011latent} at the estimated $\vp_\text{ankle}$, rotated 
% by either 90, 180, 270, or 360
in 45° steps from 0° to 360° degrees as shown in Figure~\ref{fig:initialization}, and select the rotation with the lowest reconstruction error before optimization.

\paragraph{3D pose optimization.} Using the virtual camera (\emph{Case~II} above), the optimization of the 3D pose $\vp$ under mirror constraints becomes a classical multi-view reconstruction problem. We optimize the skeleton parameterization $\vtheta$ that, when reprojecting the associated 3D joint positions $\vp$ to the real and virtual cameras, minimizes the Euclidean distance to real 2D pose $\vq$ and virtual 2D pose $\bar{\vq}$,
\begin{equation}
\textbf{}
\cL_\text{p} = \sum_t \|\vq - \Pi ( \vp(\vtheta^{(t)},\vell)) \|^2 + \|\bar{\vq} - \Pi( \mA \vp(\vtheta^{(t)},\vell)) \|^2 
\\
\label{eq:objective}
\end{equation} 
with $\vp(\vtheta^{(t)},\vell)$ the forward kinematic model and $\Pi$ the perspective projection using 
%$\mA$ and 
$\mK$. When 2D detection confidences are available, we use them as weights in Eq.~\ref{eq:objective}.

\paragraph{Smoothness and ground-plane constraints.} 
%As we currently optimize the 3D pose per frame, observing jittering effects is almost non-neglible. 
Frame-wise pose optimization leads to noisy reconstructions and inconsistent global orientations. To mitigate these, we encourage a constant velocity for location and joint angles across the video (for valid frames), referred to as \textit{location} and \textit{orientation smooth} in Eq.~\ref{eq:objective2}. To reduce floating, we utilize our ground plane estimate and constrain the lower feet by minimizing their distance to the ground plane, as described in Eq.~\ref{eq:objective2} where $\vf_{gd} = (\vm - \vf_{i})$, $\vm$ is the mirror location and $\vf_{i}$ is the closest foot (heel) to the ground. Lastly, we refine the mirror and ground normal, $\textbf{n}_{m}$ and $\textbf{n}_{g}$, during optimization and enforce both quantities to be orthogonal.

We combine all additional objectives from above as
\begin{multline}
%\begin{equation} 
\textbf{}
\cL_\text{sfo} = 
\underbrace{\lambda_{p} \|\frac{\vd^{2} \vp(\vtheta^{(t)}, \vb)}{\vd_{t}^{2}} \|}_{\text{location smooth}} +
\underbrace{\lambda_{\vtheta} \|\frac{\vd^{2}\vtheta_{k}} {\vd_{t}^{2}} \|}_{\text{orientation smooth}} + 
\underbrace{\lambda_{f} (\vn_{g} \vf_{gd})^2}_{\text{feet constraint}} \\ + 
\underbrace{(\vn_{g} \vn_{m})^2}_{\text{orthogonality}} 
+ 
\underbrace{(\|\vn_{m}\|_{2} - 1)^2}_{\text{mirror normal loss}}  + 
\underbrace{(\|\vn_{g}\|_{2} - 1)^2}_{\text{ground normal loss}},
\label{eq:objective2}
%\end{equation} 
\end{multline}
where $\frac{\vd^{2} \vp(\vtheta^{(t)}, \vb)} {\vd_{t}^{2} t} $ and $\frac{\vd^{2} \vtheta_{k}}  {\vd_{t}^{2}}$ are the second-order derivatives for the joint locations and bone orientations for all frames $t$, $\vf_{gd}$ is the vector from \danred{ground plane location} to the lower feet of interest, and $\lambda_p$, $\lambda_{\vtheta}$, $\lambda_f$ are hyper-parameters that balance the influence of the smoothness terms and the feet loss. Our final objective $\cL_\text{pose}$ is the sum of all individual terms, 
$
\cL_\text{Pose} = \cL_\text{p} + \cL_\text{sfo} .
$

\subsection{Neural Rendering and Refinement (Step 3)} \label{ssec:neural_render}

With the 3D pose $\vp(t)$ reconstructed approximately for each pair of 2D detections $\vq(t)$ and $\bar\vq(t)$ in every frame of the video, we train a generative model $G(\vtheta)$ conditioned on pose $\vtheta$. %that maps $\vp$ to $G$. 
Starting from A-NeRF~\cite{su2021nerf} %in Section~\ref{ssec: anerf_from_spin}
applied to only the real person as a baseline, we introduce our Step 2 + A-NeRF, naive mirror integration, and full mirror integration with and without efficiency-improving extensions.

\paragraph{A-NeRF initialized by Step 2 (Step 2 + A-NeRF).} % (\dan{e.g Tim model})} 
To apply articulated neural radiance fields, such as \cite{su2021nerf} and \cite{su2022danbo}, to our setting, we segment both persons in the image using \cite{lin2022robust}. We then use the real-mirror person assignment from Step 1 and Step 2 to determine the mask for the real person $\mM$ that contains the 2D keypoints associated to the real person. Our contribution is on how to also include the mirrored person and its mask $\bar{\mM}$. For the real person, we can apply existing methods, besides minor modifications to run with our skeleton definition. Input is the image $I$, skeleton $\vv^\text{ref}$, the bone lengths $\vell$ and joint angles $\vtheta$. We cast rays to the real person in the scene using pixels $(u,v)$ within the mask $\mM$, and query 64 points $\{\vb_k\}^{64}_{k=1}$ along that ray direction $\vr = \mK^{-1}(u,v,1)$
%using stratified sampling ~\cite{mildenhall2020nerf, max1995optical}
%, and determine the near and far bounds as they intersect with
within the 3D bounding box %centered at 
containing the skeleton $\vp$. 
%We sample 64 points in all experiments.
%
By using the skeleton-relative encoding from~\cite{su2021nerf} or \cite{su2022danbo}, we first map the queried points $\vb$ to the local space of each joint using $T_i$,

\begin{equation} 
\tilde{\vb_{i}} = T_i^{-1}(\vtheta_{i}, \vv_i)[\vb].
\end{equation}
A fully-connected neural network then predicts color $\gamma_{k}$ and density $\sigma_{k}$ as a function of the transformed queries, $\gamma_{k}, \sigma_{k} = \phi([\tilde{\vb}_1,\dots,\tilde{\vb}_J])$ for every sample $k$. The image is formed by volume rendering, integrating color along the ray while accounting for the transmittance computed from the density, as in the original NeRF.

The objective is the photometric loss $\cL_\text{Neural}$ between the generated and observed image, and both the joint angles and parameters of the underlying neural network are optimized jointly. Note that existing articulated NeRF models only apply as we tuned Step 2 to be compatible by introducing the additional smoothness constrained on bone orientation.

\paragraph{Layered mirror representation.}
\label{ssec: mirror-aware}

\begin{figure}[!t]%
\begin{center}%
\includegraphics[width=0.6\linewidth,trim={4.0 2.1cm 2.5 1.4cm},clip]{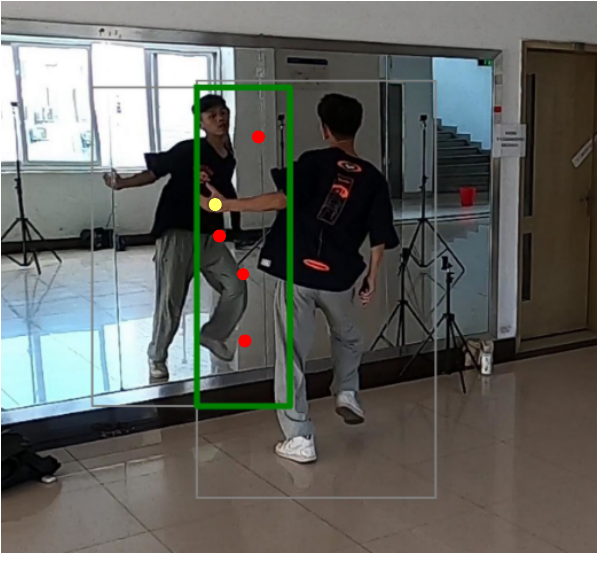}%
\end{center}%
  \dotcaption{Occlusion handling}{First, we automatically generate 
  %\danteal{automatic} 
  2D bounding boxes (in grey) from our optimized 3D keypoints. Then we shoot rays (dots) randomly in the intersection area (in green) where occlusions may happen
  %to resolve overlapping pixels 
  (in yellow).
  }
\label{fig:overlap_box} 
\end{figure}%

Mirror occlusion cases, such as the one in Figure~\ref{fig:overlap_box} where the real and mirrored person overlap, are important, as they result in errors in the segmentation masks and can lead to a few but large reconstruction errors. To make occlusion resolution part of the learning process but maintain efficiency, we automatically detect frames where occlusion occurs by measuring the intersection over union (IOU) of the bounding boxes $\mN$ and $\bar{\mN}$ enclosing the projected real and mirrored 3D poses from Section~\ref{ssec:poselift}. 

Given these boxes, we compute an intersection box that bounds overlapping areas and shoot rays randomly within the intersection box to resolve occluding pixels. Since each pixel is at an intersection of $\mN$ and $\bar{\mN}$, we process the occlusion samples along the direct view ray, 
$\{{\vb}_k\}_{k=1}^{64}$, and along its mirrored path, $\{\bar{\vb}_k\}_{k=1}^{64}$. 
\emph{Case~II} gives the reflected view-ray %$\bar{\vr}$
\begin{equation}
\bar{\vr} = \mA_{3\times3}\vr %\quad
\label{eq:mirror_naive},
\end{equation} 
with the origin at virtual camera center $\bar{\vc}$.
%\hr{with origin at the intersection of the original ray with the mirror.} 
Note that we do not bound the occlusion samples to $\mM$ and $\bar{\mM}$ as the segmentation masks are often unreliable when people overlap.

Furthermore, sampling a different number of samples for occlusion rays does not fare well with the batch processing in NeRF. To make it compatible, we process real and mirror samples, including occlusion cases, independently to yield image layers $\mL$ and $\bar{\mL}$ and corresponding alpha maps $\alpha$ and $\bar{\alpha}$. We exploit that if occlusion happens, the real person occludes the mirrored person. This holds in general, since the mirror image results from an indirect light path that is always longer than the direct view, and enables
combining these partial results using back-to-front layering. Starting from the background image $\mI_\text{bg}$, the final image is
\begin{gather} 
\hat{\mI} =  \mL  \alpha + 
(1-{\alpha}) 
\left(\bar{\mL}  \bar{\alpha} + (1-\bar{\alpha}) I_\text{bg}
\right).
\end{gather} 
This layered composition enables efficient training on $\cL_\text{Neural}$ and rendering in batches of the same size, while %properly
accounting for mirror occlusion.

\begin{figure*}[!t]
\centering
\includegraphics[width=0.82\linewidth,trim={0.0cm 0.0 0.0cm 0.0cm},clip]{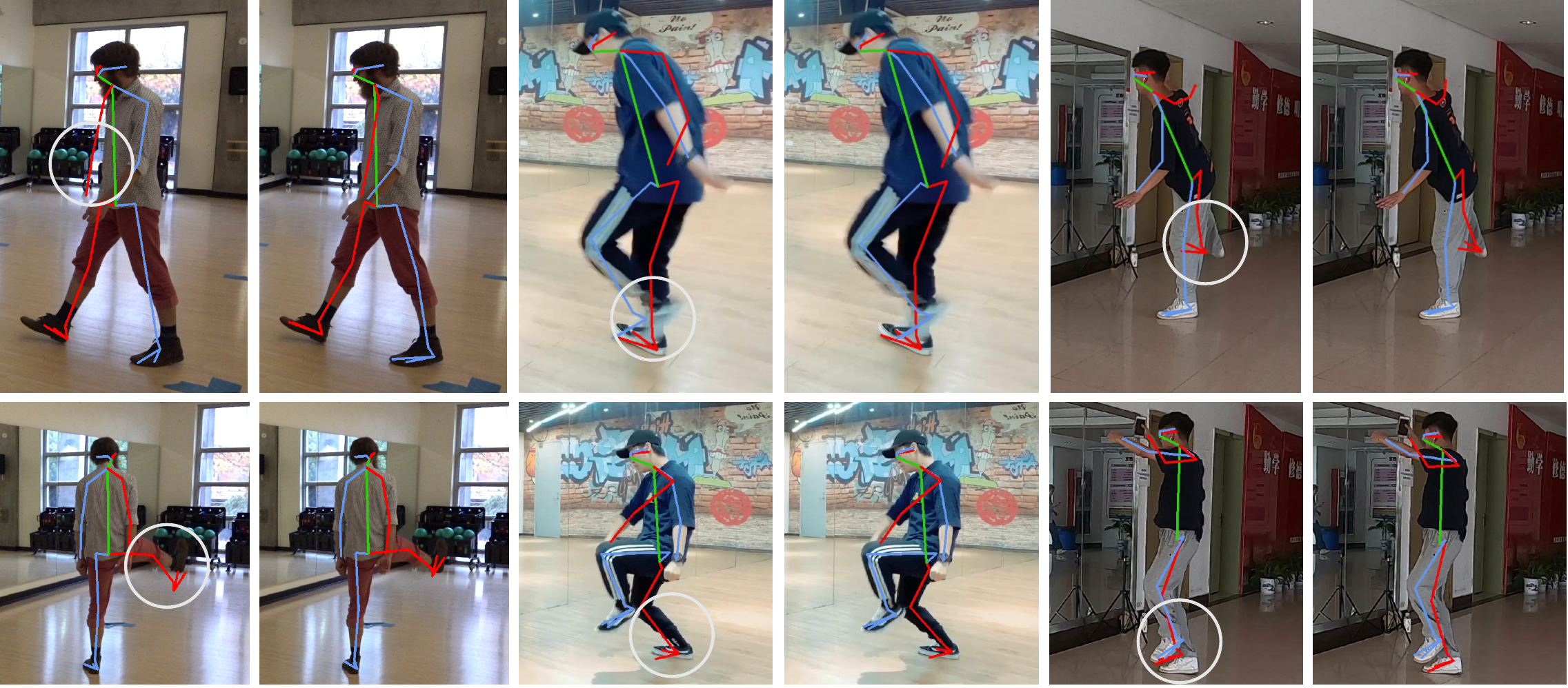}
  \caption{\textbf{Qualitative results of pose refinement on a new diverse sequence, internet video, and \emph{MirrorHuman-eval} dataset~\cite{fang2021reconstructing} \da{respectively}}. Our Mirror A-NeRF aligns the skeleton model well to the images.  \textbf{Left:} Before refinement (after Step 2) \textbf{Right:} After the volumetric model refinement (Step 3). 
  }
\label{fig:before and after} 
\end{figure*}

\paragraph{Baseline w/o layering.}
\label{ssec:mirror_bpsa}

Without the introduced layering, it would require to sample twice as many points over the union of both masks $\mM$ and $\bar{\mM}$, the samples $\{\vb_k\}_{k=1}^{64}$ along the ray up to the mirror and samples $\{\bar{\vb}_k\}_{k=1}^{64}$ after the mirror reflection. However, unless real and mirrored person occlude, one of the two sample sets is in empty space, thereby training NeRF to merely predict $0$ density for that set, 
\danred{leading to slow}
% leading to extremely slow 
convergence.  %of the NeRF
%and low-quality models in our preliminary experiments.

\paragraph{Baseline w/o occlusion}
Assuming that the person masks $\mM$ and $\bar{\mM}$ are correct and non-overlapping, one can render two separate images of real and mirror person by respectively sampling $\mM$ and $\bar{\mM}$, using \emph{Case~II} with the notion of two cameras.
By minimizing the photometric objective, this baseline is efficient and utilizes the information across both views but does not account for mirror occlusions and imperfect masks around occlusion areas.

\begin{table*}[!ht]
\centering
  \resizebox{0.65\linewidth}{!}{%
  \begin{tabular}{|l|ccc|c|}
%\hline
%\multicolumn{4}{c}{} & {PA-MPJPE $\downarrow$}
% & \multicolumn{7}{c}{PA-MPJPE $\downarrow$}  \\
\hline
Method                                 & 3D Training & Mirror Calibration & 2D Input & PA-MPJPE $\downarrow$    \\
\hline
SMPLify-X \cite{pavlakos2019expressive}      &  \textcolor{Orange}{3D pose prior} &  n/a &  \textcolor{OliveGreen}{detections}  & 90.57   \\
A-NeRF \cite{su2021nerf}                         &  \textcolor{Orange}{partial (init.)} & n/a  &   \textcolor{OliveGreen}{detections} & 84.72 \\  % incorrect 84.70     \\
SPIN \cite{kolotouros2019spin}                         &  \textcolor{Red}{supervised} & n/a  &  \textcolor{OliveGreen}{detections}   &   67.42     \\
SPIN \cite{kolotouros2019spin}+SMPLify \cite{pavlakos2019expressive}         & \textcolor{Orange}{partial (init.)} & n/a & \textcolor{OliveGreen}{detections}  &  \textbf{61.47}     \\
\hline
%Ours (detections)                      & 225.7 & 52.1 & 129.0 & 64.7 & 76.2 & 141.5 & 112.4 \\
%Ours (GT 2D pose input)                & 81.3 & 43.7 & 112.0 & 47.19 & 47.05 & 75.18 & \textbf{61.3}\\
Ours (Step 2)                     &  \textcolor{OliveGreen}{unsupervised} & \textcolor{OliveGreen}{automatic} &  \textcolor{OliveGreen}{detections} &  63.00\\ %59.00\\
Ours (Step 2 + A-NeRF \cite{su2021nerf})                     & \textcolor{OliveGreen}{unsupervised}   & \textcolor{OliveGreen}{automatic} &  \textcolor{OliveGreen}{detections} & 62.69\\ 
Ours (Step 3, w/o occlusion)                      & \textcolor{OliveGreen}{unsupervised} & \textcolor{OliveGreen}{automatic} &  \textcolor{OliveGreen}{detections} & 61.46\\ %\textbf{57.73 }
Ours (Step 3, w occlusion)                      & \textcolor{OliveGreen}{unsupervised} & \textcolor{OliveGreen}{automatic} &  \textcolor{OliveGreen}{detections} & \textbf{61.30}\\
\hline
Ours (Step 2, using GT input)                     &   \textcolor{OliveGreen}{unsupervised} & \textcolor{OliveGreen}{automatic} & \textcolor{Red}{manual}  & 39.53 \\ %59.00\\
MirrorHuman \cite{fang2021reconstructing} (w/o mirror GT)   & \textcolor{Orange}{partial (init.)} & \textcolor{OliveGreen}{automatic}  & \textcolor{Red}{manual} &  33.24     \\
MirrorHuman \cite{fang2021reconstructing} & \textcolor{Orange}{partial (init.)} & \textcolor{Red}{manual}   & \textcolor{Red}{manual}    & \textbf{32.96}     \\
%DMB \cite{liulearning}  & - & -    & - & - & -  \textcolor{OliveGreen}{XXX}     \\
\hline
%\textbf{57.73 }
% Ours (detections - previous)                      & 218.0 & 56.4 & 216.7 & 77.9 & 91.8 & 138.83 & 146.1 & 75.2 \\
% Ours (GT 2D pose input)                & 77.4 & 43.7 & 104.8 & 43.9 & 38.9 & 71.8 & \textbf{57.4} & \textbf{42.0}\\
\end{tabular}
%\tiny *using 14 joints as no nose keypoint included
}
\dotcaption{3D pose reconstruction} {Ours is the only fully-automatic mirror-based method. We match the accuracy of off-the-shelf 3D pose estimators with only 2D detections as input and reproduce results of mirror methods using GT input~\cite{fang2021reconstructing}.
}
% 
% and two variants of the hybrid approach \cite{fang2021reconstructing} (middle)
\label{tab:3D comparison}
\end{table*}

\section{Evaluation}
\label{sec:evaluation}

We compare our Mirror-aware Neural Human on the tasks of body pose and appearance reconstruction against state-of-the art methods to verify that existing NeRF-based body models benefit largely from the additional geometric mirror constraints, that every stage of our motion capture pipeline brings improvement, and ablate model choices. The supplemental document and video provide additional examples.

\parag{Variants and Baselines.}
We integrated our \emph{Mirror-aware Neural Human} formulation into A-NeRF~\cite{su2021nerf} and DANBO~\cite{su2022danbo} and refer to them as \emph{Mirror A-NeRF} and \emph{Mirror DANBO}. We use the original A-NeRF~\cite{su2021nerf} and DANBO~\cite{su2022danbo} as baselines, as well as the mirror-based pose estimation method, MirrorHuman~\cite{fang2021reconstructing} and the established single-view reconstruction techniques SPIN \cite{kolotouros2019spin} and SMPLify~\cite{pavlakos2019expressive}. For pose estimation accuracy, we only compare to the A-NeRF neural body method as DANBO does not support pose refinement.

\parag{Benchmark Datasets.} We use the \emph{MirrorHuman-eval} dataset from \cite{fang2021reconstructing}.
It contains a complex dancing performance in front of a large mirror and is captured with six cameras arranged in an arc around the mirror. We exclude camera 4 as it is directly frontal to the mirror and leads to degenerate configuration where the mirrored person is largely occluded. Since no validation set was specified, we use camera 2 and 3, that are placed between a 45-to-90 degree angle to the mirror, for tuning hyperparameters and test on all cameras as previous methods did. Following~\cite{fang2021reconstructing}, we optimize the pose separately for each video, treating this dataset as five independent recordings instead of a multi-view setup. We %follow \cite{fang2021reconstructing} 
also concur in evaluating \da{pose accuracy for} Step 2 and 3 on every 100th frame while reconstructing in-between frames only to ensure smoothness. For the neural models, \da{we quantify image reconstruction accuracy} on every 20th frame and withhold the last 10\% of frames to test novel pose synthesis.

\parag{Additional qualitative sequences.} To be able to demonstrate generality and showcase the simplicity of capturing with a mirror, we utilize the internet dancing recordings from \cite{fang2021reconstructing} and recorded a new dataset that is more diverse in terms of appearance, e.g., including a beard, male and female, loose clothing, and casual everyday motions that go beyond dancing (see Figure~\ref{fig:before and after}). We used a single camera for recording and employed 2000 frames for reconstruction.

\parag{Metrics.} For pose estimation tasks, we report the scale-normalized MPJPE (N-MPJPE) introduced in~\cite{rhodin2018unsupervised}, and the Procrustes-aligned MPJPE (PA-MPJPE), both in mm over the 15 joints defined in~\cite{fang2021reconstructing}. We omit MPJPE without scale normalization as monocular reconstruction is inherently scale-ambiguous~\cite{gunel2019face}. For image synthesis, we quantify the image quality by PSNR and SSIM~\cite{wang2004ssim}.

\parag{Implementation details.} In Step 2, we optimize 3D poses for 2K iterations. For Step 3, we train the neural rendering model up to a maximum of 300K steps for DANBO and a maximum of $2\times200$K for A-NeRF with pose refinement and appearance fine-tuning.

\begin{table}[t]
\centering
  \resizebox{0.98\linewidth}{!}{%
\begin{tabular}{|l|cc|}
\hline
Method & Cam 6 PSNR $\uparrow$ & Cam 6 SSIM $\uparrow$ \\
\hline
A-NeRF~\cite{su2021nerf} &  25.52 & 0.8662 \\
% old without offset: 25.67, 0.9810
Ours(Mirror A-NeRF w/o Occlusion)  & \cellcolor{lightgray!25} \textbf {25.89} &  \cellcolor{lightgray!25} \textbf {0.9210}\\
DANBO~\cite{su2022danbo}  & 28.97 & 0.9193 \\
Ours(Mirror DANBO w/o Occlusion)  & \cellcolor{lightgray!25} \textbf {31.87} & \cellcolor{lightgray!25} \textbf {0.9522} \\
\hline
\end{tabular}
}
\shortcaption{Quantitative image reconstruction accuracy} {on \emph{MirrorHuman-eval} dataset in terms of PSNR and SSIM. A-NeRF~\cite{su2021nerf} and DANBO~\cite{su2022danbo} without mirror remain blurry, leading to lower scores. }
\label{tab:novel_view_generalization}
\end{table}

\subsection{Mirror Normal Estimation}

Our average normal estimation error (using 2D detections as input) is 0.4° compared to the GT normal provided in~\cite{fang2021reconstructing}. Camera 4 is excluded as the real person occludes the mirror image in most frames. This automatic mirror calibration is highly accurate and very close to the 0.5°  obtained from the vanishing point method in~\cite{fang2021reconstructing} on the same cameras.

\subsection{Pose Estimation}
\paragraph{Comparison to refining pose with a dense body model.} 
%Figure~\ref{fig:pose_refinement_visual} shows qualitative results and 
Table~\ref{tab:3D comparison} shows the results for the MirrorHuman-eval dataset.  Compared to A-NeRF~\cite{su2021nerf}, which also refines SPIN estimates using a volumetric body model, our method improves 3D pose estimates significantly, by more than $20 \%$. This highlights the importance of integrating a second view for accurate reconstruction, here via the mirror. Figure~\ref{fig:before and after} shows that, by benefiting from the NeRF model, the joint refinement of pose and shape improves significantly, particularly on extreme poses.

\parag{Comparison to methods lifting 2D pose to 3D.} %Table~\ref{tab:3D comparison} shows the results for the MirrorHuman-eval dataset. 
We outperform the supervised approach SPIN \cite{kolotouros2019spin} and single-view optimization SMPLify-X \cite{pavlakos2019expressive}, and match their combination, %as these do not generalize well to a new dataset without retraining on 3D pose labels.
as these single-view approaches do not fare well under occlusion and are prone to depth ambiguity.

\parag{Comparison to existing mirror approaches.}
The prior method \cite{fang2021reconstructing} focuses on controlled conditions, using manually corrected 2D ground-truth (GT) and a pre-trained pose estimator for initialization. To compare on fair grounds, we run a variant that also uses 2D GT as input. Table~\ref{tab:3D comparison} shows that it matches the accuracy up to 7mm. The remaining discrepancy can be attributed to our method not being tuned for GT input and not using a 3D pose estimator, which are known to not generalize well to very extreme motions.  The other existing mirror work \cite{liulearning} cannot be compared to as it does not provide joint angles for shape reconstruction and does not evaluate on publicly available datasets.

\begin{figure*}[t]
\centering
\includegraphics[width=0.95\linewidth,trim={0.0cm 0.0 0.0cm 0.0cm},clip]{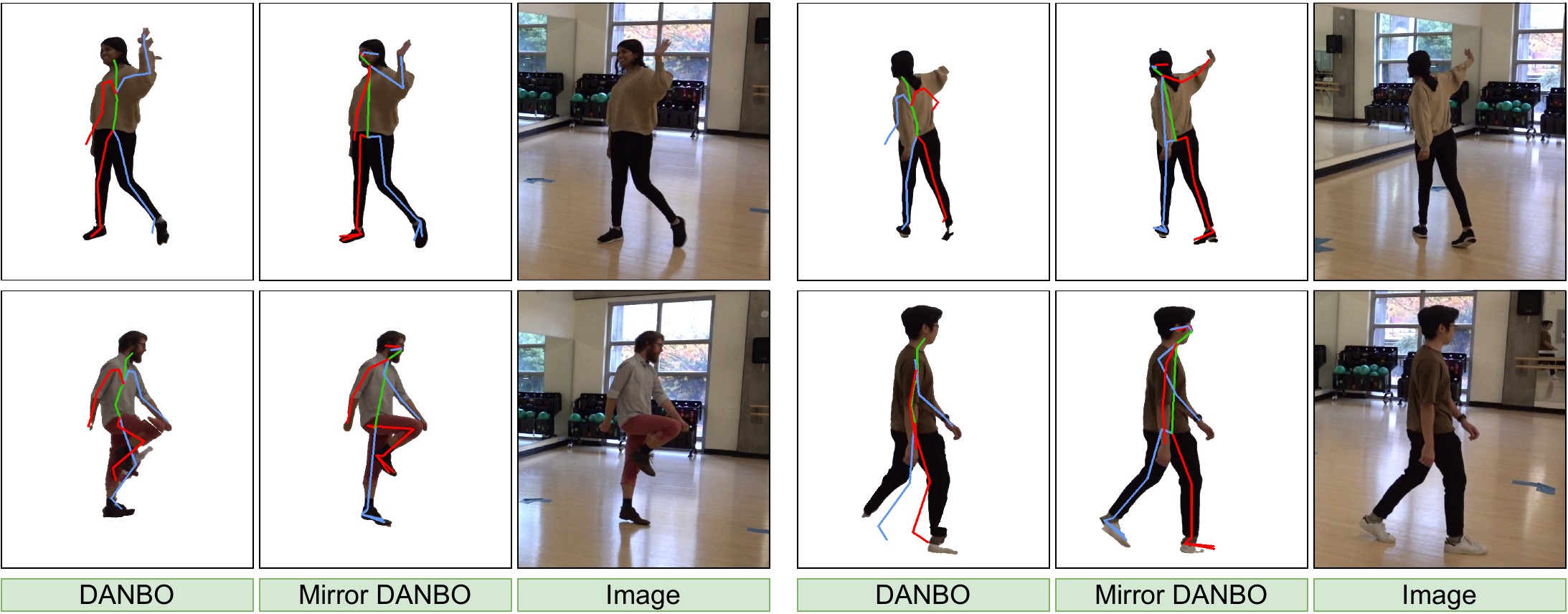}
  \caption{\textbf{Neural body reconstruction} on three different subjects from the additional qualitative sequences, including loose clothing, challenging poses, and casual day-to-day motion. 
  Our {\modelname} reconstructs the body details considerably benefiting from the mirror model and skeleton.
  }
\label{fig:qualitative} 
\end{figure*}

\subsection{Body Shape and Appearance}
Figure~\ref{fig:top_figure} and Figure~\ref{fig:qualitative} show %novel view 
the images synthesized by different neural rendering models. For better visualization, we apply connected component analysis to remove 
% small 
floating artefacts stemming from shadowing and background.
%As opposed to classical mirror reconstruction with stereo cues, these neural rendering models recover occluded parts by learning a single model for the video frames, relative to the underlying skeleton pose. 
%
On \emph{MirrorHuman-eval dataset~\cite{fang2021reconstructing}}, both of our variants, Mirror A-NeRF and Mirror DANBO, synthesize sharper results with more fine-grained details
%less floating artifacts 
compared to the mirror-less counterparts. We attribute the improvement to the better pose initialization from Step 2 and the additional appearance supervision by the mirror view, all enabled by our mirror modeling. Table~\ref{tab:novel_view_generalization}
validates the visual improvement quantitatively in terms of image reconstruction accuracy.
%novel view synthesis on held-out camera views. 
Both Mirror A-NeRF and Mirror DANBO better learn the body shape and appearance, verifying the effectiveness of our \emph{Mirror-aware Neural Humans}.

\begin{table}[b]
\centering
  \resizebox{0.98\linewidth}{!}{
\begin{tabular}{|l|c|c|}
% \begin{tblr}{
%     colspec = {l|c|c},
%     % row{2} = {purple7},
%     % column{2} = {teal7},
%     cell{0}{0} = {yellow7},
%   }
\hline
Components removed from model & N-MPJPE $\downarrow$        & PA-MPJPE   $\downarrow$    \\
% $\theta^{(t)} + \vh^{(t)}$  & \phantom{0}91.3          & 76.7        \\
% $\theta^{(t)} + \vh^{(t)} + \vb$ & \phantom{0}55.1          &    48.2           \\
% $\theta^{(t)} + \vh^{(t)} + \mA$             & 127.0         &  93.0\\
% $\theta^{(t)} + \vh^{(t)} + \mA(\vn_m,\vm)$        & \phantom{0}96.2          & 74.5 \\
% $\theta^{(t)} + \vh^{(t)} + \mK$             & 100.3         & 78.0       \\
% $\theta^{(t)} + \vh^{(t)} + \vb + \mA$    & \phantom{0}57.4          & 53.4          \\
% $\theta^{(t)} + \vh^{(t)} + \vb + \mA(\vn_m,\vm)$    & \phantom{0}{\ul51.9}    & {\ul48.2}\\
% $\theta^{(t)} + \vh^{(t)} + \mK + \vb$         & \phantom{0}{\ul\textbf{50.8}} & {\ul\textbf{47.2}} \\
% $\theta^{(t)} + \vh^{(t)} + \mK + \mA$          & 128.9         & 92.9         \\
% $\theta^{(t)} + \vh^{(t)} + \mK + \mA(\vn_m,\vm)$     & \phantom{0}95.9          & 77.1     \\
% $\theta^{(t)} + \vh^{(t)} + \mK + \vb + \mA$      & \phantom{0}55.1          & 51.3       \\
% $\theta^{(t)} + \vh^{(t)} + \mK + \vb + \mA(\vn_m, \vm)$ & \phantom{0}{\ul51.1}    & {\ul47.9} \\
\hline
% Base (w/o smoothness and feet constraint)     & \textcolor{red}{\phantom{0}98.34}          & \textcolor{red}{62.46}     \\

Base w/o smooth.~and feet constr.     & \phantom{0}99.94       & 63.03     \\
% \hline
% $\theta^{(t)} + \vh^{(t)} + \mK + \vb + \mA$      & \phantom{0}55.1          & 51.3       \\
w/o location smooth.      & \phantom{0}93.79          & 57.98     \\
% $\theta^{(t)} + \vh^{(t)} + \mK + \vb + \mA$      & \phantom{0}55.1          & 51.3       \\
% \hline
w/o orientation smooth.     & \phantom{0}92.88          & 57.93    \\
% $\theta^{(t)} + \vh^{(t)} + \mK + \vb + \mA$      & \phantom{0}55.1          & 51.3       \\
w/o feet constraint  & \cellcolor{lightgray!25}\phantom{0}{84.73}    & \cellcolor{lightgray!25}{54.06} \\
\hline
Full objective  & \cellcolor{gray!25}\phantom{0}{\textbf{62.45}}    &  \cellcolor{gray!25}{\textbf{44.15}}\\
% \end{tblr}
\hline
\end{tabular}}
\shortcaption{Ablation study} {on the optimal regularization configuration to reduce the influence of noisy detections. All our contributions improve on the baseline.
}
\label{tab:ablation}
\end{table}
\begin{table}%[t]
\centering
  \resizebox{0.98\linewidth}{!}{%
\begin{tabular}{|l|cc|}
\hline
Method & Cam 6 PA-MPJPE   $\downarrow$ & Cam 7 PA-MPJPE   $\downarrow$  \\
\hline
A-NeRF~\cite{su2021nerf} & 86.21   & 89.10  \\
%A-NeRF~\cite{su2021nerf} & \dan{85.11}   & \dan{90.06}  \\
Ours (Step 2 + A-NeRF) & 51.21   & 58.61  \\
Ours (Mirror A-NeRF)  & 48.84 & 57.82  \\
Ours (Mirror A-NeRF w/ occlusion)  & \textbf{48.51} & \textbf{57.32}  \\
\hline
\end{tabular}}
\dotcaption{Pose refinement on \emph{MirrorHuman-eval} dataset~\cite{fang2017rmpe} on videos with occlusions} {With occlusion handling, \modelname{} improves upon A-NeRF~\cite{su2021nerf}.} 
\label{tab:pose_refinement_quant_nos}
\end{table}

\subsection{Ablation Study}
To analyze the effect of our model choices in Step 2, we use camera 3
%with the same subsampling of every 100th frame and 
in the \emph{MirrorHuman-eval} dataset
%MirrorHuman 
over all 19 joints that have ground truth.
We perform experiments with different configurations for the joint optimization of the bone factors, global position, and mirror parameters. Table~\ref{tab:ablation} validates that each component contributes to the final result. The effect of and robustness to different weights of the smoothness terms are evaluated and explained in the supplemental material. 

We also analyze the different ways of integrating mirror constraints and occlusion handling on cameras 6 and 7. Table~\ref{tab:pose_refinement_quant_nos} presents the pose refinement outcomes after the neural rendering steps (Section~\ref{ssec:neural_render}). Running A-NeRF with our Step 2 poses already improves due to the more accurate pose initialization. 
%Mirror-Aware Neural Human 
Our Mirror A-NeRF further improves, as it enables pixel-level multi-view (real and mirror pose)  refinement. Taking occlusion into account further surpasses the two baselines. Note that the total improvement of handling occlusions computed over all frames and all joints is small and we therefore only \danred{enable it on sequences that include a} sufficient number of occlusion frames.
\section{Limitations and Future Work} 
\label{sec:limitations}
Our 3D reconstruction algorithm only works in cases where the person's mirror image is visible most of the time, restricting the camera placement close to 20-to-70 degrees to the mirror.
When the camera view direction is close to parallel or orthogonal to the mirror, 3D triangulation is unreliable. 
%Hence, outliers cannot be detected or corrected. 
Moreover, the 3D reconstruction is sensitive to the initial reconstruction of the ground plane and focal length, e.g., errors emerging when bystanders with varying heights violate the constant height assumption.

Moreover, since we rely on an estimated segmentation mask, sometimes body parts are cut out or the shadow or other background parts in the scene are represented in the 3D volumetric model \da{creating opportunities for future work}.
In the future, we will attempt to apply similar techniques to animal motion capture, which, however, requires redefining our upright standing assumption and filtering in Step~1. 

\section{Conclusion} 

Our method reconstructs a 3D neural body model from mirror images by treating the mirror as a second camera and calibrating the camera extrinsics and mirror geometry directly from the motion of people detected in 2D. 
%Subsequently, we use this geometric human body model as initialization to learn and refine a neural human body model without requiring any ground-truth data or 3D scan.
This alleviates manual \da{mirror} annotation and initializing with pre-trained models, 
%compared to other human mirror-based approaches and 
which lets us reconstruct difficult and complex human performances for which existing approaches struggle. 

% Mirror-aware Neural Humans
\modelname{} let anyone with a mirror and camera reconstruct a full 3D human model.
In particular, we foresee low-cost medical applications, such as mirror-based pose estimation for rehabilitation~\cite{lee2011legs}.

{
    \small
    \bibliographystyle{ieeenat_fullname}
    \bibliography{main}
}

\end{document}